\pdfoutput=1
\documentclass[preprint,10pt]{elsarticle}

\usepackage{amsmath,amssymb}
\usepackage{graphicx}
\usepackage{booktabs}
\usepackage{tabularx}
\usepackage{multirow}
\usepackage{makecell}
\usepackage{array}
\usepackage{xcolor}
\usepackage[hidelinks]{hyperref}
\usepackage{enumitem}
\usepackage{textcomp}
\usepackage[a4paper,margin=2.5cm]{geometry}
\usepackage{placeins}
\graphicspath{{figures/}}
\definecolor{rdH}{HTML}{CDEFE0}\definecolor{rdO}{HTML}{F8CFCE}\definecolor{rdU}{HTML}{FBE6B5}\definecolor{rdW}{HTML}{CFE0F6}

\newcommand{\RS}{RobustSeiz}
\newcommand{\Se}{\mathrm{Se}}
\newcommand{\PPV}{\mathrm{PPV}}
\newcommand{\Fone}{\mathrm{F1}}
\newcommand{\FPd}{\mathrm{FP/24\,h}}
\newcolumntype{L}[1]{>{\raggedright\arraybackslash}p{#1}}
\journal{Computer Methods and Programs in Biomedicine}

\begin{document}

\begin{frontmatter}

\title{\RS{}: a model-agnostic framework for auditing the recording-\allowbreak environment, noise and adversarial robustness of EEG seizure detectors}

\author[sut]{Mohammad Mohammadi\corref{cor1}}
\ead{mohammadm97i@gmail.com}
\author[sut]{Alireza Zarei}
\cortext[cor1]{Corresponding author.}
\affiliation[sut]{organization={Department of Mathematical Sciences, Sharif University of Technology},
            addressline={Azadi Ave., P.O. Box 11365-8639},
            city={Tehran},
            country={Iran}}

\begin{abstract}
\noindent\textbf{Background and Objective:} Automated electroencephalography (EEG) seizure detectors are validated on clean research recordings, whereas clinical recordings are altered by electrode problems, equipment settings and artifacts. As seizure treatment is time-critical, we tested whether detectors stay reliable and timely under such conditions.

\noindent\textbf{Methods:} We developed \RS{}, an open-source framework that audits any trained detector without modifying it. Its novelty is the evaluation pipeline, not a new detector: a model-agnostic protocol applying controlled, seeded shifts (recording environment, noise and artifacts, worst-case adversarial), scoring that adds onset timing and model uncertainty to event-based accuracy, and paired statistics against clean recordings. We audited four deep-learning detectors on three public datasets unseen in training.

\noindent\textbf{Results:} Robustness varied widely and was independent of clean accuracy. Adversarial changes of 5--10\% of signal amplitude reduced the F1-score by up to 0.57, and electrode-contact noise silently lowered sensitivity from 0.57 to 0.06. Shifts delayed the seizures still detected by up to 62\,s. Read with accuracy, dropout certainty fell under filtering shifts but rose as noise silenced detection; only a quarter of failures lowered it.

\noindent\textbf{Conclusions:} Clean accuracy alone does not establish reliability or timeliness. \RS{} makes pre-deployment robustness audits reproducible.
\end{abstract}


\end{frontmatter}

\section{Introduction}
\label{sec:intro}

Scalp electroencephalography (EEG) is the reference investigation for epilepsy, which affects about 50 million people~\cite{guekht2021who,fisher2014ilae,peltola2023ifcn}. Automated seizure detection promises to relieve expert review in monitoring units and at home~\cite{ryvlin2018homeeeg,beniczky2021automated,koren2021commercial}, and deep detectors trained on large public corpora report strong accuracy~\cite{shah2018tusz,schirrmeister2017deep,wu2025seizuretransformer,seeuws2024eventnet}. Few have reached routine clinical use~\cite{moutonnet2024clinical,koren2021commercial}.

Part of the gap is methodological. The Seizure Community Open-source Research Evaluation framework (SzCORE) made accuracy comparable by fixing corpora in a common BIDS-EEG format, evaluation splits and clinically motivated event scores~\cite{dan2024szcore}. Its first challenge then scored 28 algorithms on 4360\,h of unseen clinical EEG: the best event F1 was 0.32~\cite{dan2025challenge}. Clean accuracy on a research corpus predicts accuracy in a new setting poorly.

Clinical EEG departs from curated corpora in ways that are known in advance. Electrode impedance drifts and contacts fail, amplifiers differ in filters and sampling rates, mains interference and notch settings depend on the country, and movement produces impulsive and broadband artifacts~\cite{tatum2018artifact,uriguen2015artifact,kappenman2010impedance}. Each is a structured \emph{dataset shift} under which a model's errors can grow without warning~\cite{quinonero2009shift,finlayson2021datasetshift}. Adversarial machine learning adds that small, deliberately chosen input changes can flip the outputs of medical and physiological models~\cite{goodfellow2015fgsm,madry2018pgd,finlayson2019adversarial,wu2023adversarialreview}; even where attacks are not a realistic threat, the smallest harmful perturbation measures how close decisions lie to the decision boundary.

A detector that misses seizures under poor contact fails \emph{silently}; one that floods the ward with false alarms fails \emph{loudly} and is switched off; one that alarms late costs treatment time. A clean F1 score sees none of these. Reporting guidelines now ask for robustness and failure analysis~\cite{collins2024tripodai,mitchell2019modelcards}, and controlled corruptions have exposed such failures reproducibly in computer vision~\cite{hendrycks2019corruptions,koh2021wilds}. EEG perturbation studies have so far examined learned representations and classification tasks~\cite{wagh2022latent,sirca2026beyond,lee2026tta,kastrati2025eegbench}. An end-to-end protocol that stresses complete seizure \emph{detectors}, including their own preprocessing and event post-processing, with clinical event metrics, has been missing.

We present \RS{}, which adds a robustness dimension to the SzCORE standard. Its novelty lies in the evaluation methodology and its software rather than in a new detector or attack. We contribute (i) a formal, model-agnostic protocol that applies seeded perturbations in memory to read-only held-out data, outside the detector, with subject-level paired confidence intervals (Sections~\ref{sec:formal}, \ref{sec:stats}); (ii) 36 perturbation settings in environment, noise and adversarial families, each an explicit signal model (Section~\ref{sec:shifts}); (iii) an adapter interface and exact montage conversion with explicit validity rules (Sections~\ref{sec:data}, \ref{sec:models}); (iv) a report that reads event detection, onset timing and Monte Carlo dropout certainty~\cite{gal2016dropout} together (Sections~\ref{sec:scoring}, \ref{sec:mcd}); and (v) the first multi-detector, multi-corpus robustness audit of seizure detectors, with mechanistic explanations of its failure modes (Section~\ref{sec:results}). Code, adapters and all results are open.

\section{Methods}
\label{sec:methods}

\begin{figure*}[!htbp]
\centering
\includegraphics[width=\textwidth]{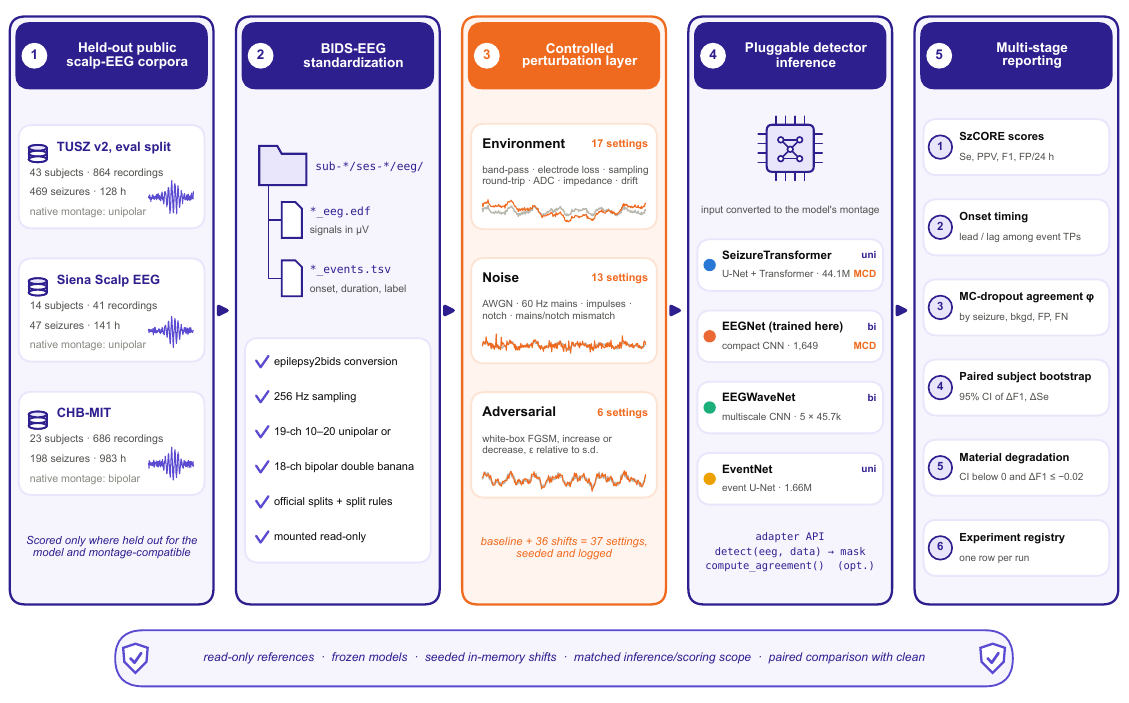}
\caption{\RS{} framework. (1) Public scalp-EEG corpora are used only where they are held out for the detector and compatible with its input montage. (2) Recordings and annotations are standardized to BIDS-EEG and mounted read-only. (3) One of 37 settings (clean baseline or one of 36 seeded shifts) is applied in memory, before any detector preprocessing. (4) The shifted recording is converted to the detector's montage and passed to the frozen detector through a two-function adapter. (5) Outputs are scored with SzCORE metrics, onset delay and Monte Carlo dropout (MCD) agreement; a paired subject bootstrap gives confidence intervals, and each run is a row of the experiment registry.}
\label{fig:framework}
\end{figure*}

Fig.~\ref{fig:framework} summarizes the framework: held-out corpora are standardized and mounted read-only, a perturbation is applied in memory, the result is converted to the detector's montage and passed to the frozen detector, and the output is scored and logged.

\subsection{Problem formulation}
\label{sec:formal}

Let a recording be $x \in \mathbb{R}^{C \times N}$ ($C$ channels, $N$ samples at sampling rate $f_s$) with reference seizure intervals $\mathcal{E}=\{(a_k,b_k)\}$. A detector is the composition
\begin{equation}
f = \pi \circ h \circ \rho
\label{eq:detector}
\end{equation}
of its own preprocessing $\rho$ (filtering, resampling, normalization, windowing), a network $h$ and post-processing $\pi$ (thresholding, event formation) that returns a binary seizure mask; $f$ is frozen. A \emph{shift} $T_{\theta}(\,\cdot\,;\xi)$, with severity $\theta$ and seed $\xi$, acts on the recording as acquired, before $\rho$, so the detector meets it as it would meet a shifted acquisition. For a held-out corpus $\mathcal{D}$ and a metric $M$ pooled over its recordings, the \emph{robustness profile} of $f$ is
\begin{equation}
\Delta M(\theta) = M\big(f\circ T_\theta;\mathcal{D}\big) - M\big(f;\mathcal{D}\big),
\label{eq:profile}
\end{equation}
evaluated on a grid of $\theta$ for each shift family. Clean and shifted runs share recordings, references and scoring scope, so $\Delta M$ is a paired, within-subject contrast. Because $\Fone\ge0$, $\Delta\Fone\ge-\Fone(f;\mathcal{D})$: a detector with low clean F1 cannot lose much, and small losses of a poor detector are a \emph{floor effect}, not robustness.

\subsection{Datasets and held-out validity}
\label{sec:data}

We used the three public scalp-EEG corpora of the SzCORE benchmark that remain obtainable (Table~\ref{tab:coverage}): the evaluation split of the Temple University Hospital EEG Seizure Corpus (TUSZ, v2)~\cite{shah2018tusz,obeid2016tuh}, the Siena Scalp EEG Database~\cite{detti2020siena} and the pediatric CHB-MIT database, recorded in a bipolar montage~\cite{shoeb2009chbmit,goldberger2000physionet}. SeizeIT1 is no longer shared, and SeizeIT2 uses behind-the-ear electrodes~\cite{bhagubai2025seizeit2}. All corpora were converted to BIDS-EEG~\cite{pernet2019bidseeg}: EDF signals~\cite{kemp2003edfplus} at 256\,Hz, 19 common-average channels of the 10--20 system (TUSZ, Siena) or the 18 double-banana derivations (CHB-MIT).

A detector--corpus pair is \emph{valid} only if (i) no recording of the corpus was used to train, validate or tune the detector, and (ii) the corpus converts exactly to the detector's input montage. The double banana is $B=Ax$, with each row of $A\in\{0,\pm1\}^{18\times19}$ the difference of two electrodes, so unipolar-to-bipolar conversion is exact. The electrode graph has three connected components, hence $\operatorname{rank}(A)=16$; even under a common-average constraint, two degrees of freedom remain, and bipolar recordings cannot be converted back (proof in Supplementary Section~S2). Unipolar detectors are therefore excluded from CHB-MIT. Both conditions give the eight valid pairs of Table~\ref{tab:coverage}.

\begin{table*}[!htbp]
\centering
\caption{Held-out corpora, detectors and valid detector--corpus pairs. Hours and seizure counts are measured on our BIDS conversions; reference events are counted after SzCORE pre-processing (events closer than 90\,s merged, events longer than 300\,s split). \checkmark: valid pair; $\times$: excluded.}
\label{tab:coverage}
\footnotesize
\resizebox{\textwidth}{!}{%
\begin{tabular}{@{}l l r r r c c c c c@{}}
\toprule
Corpus & Native montage & Subj. & Rec. & Hours & \makecell{Seizures\\(ref.\ events)} & \makecell{Seizure-\\Transformer} & EEGNet & EEGWaveNet & EventNet \\
\midrule
TUSZ v2, \texttt{eval} & 19-ch unipolar & 43 & 864 & 128 & 469 (340) & \checkmark & \checkmark & \checkmark & \checkmark$^{c}$ \\
Siena & 19-ch unipolar & 14 & 41 & 141 & 47 (47) & $\times^{a}$ & \checkmark & \checkmark & \checkmark \\
CHB-MIT & 18-ch bipolar & 23$^{d}$ & 686 & 983 & 198 (200) & $\times^{b}$ & \checkmark & $\times^{a}$ & $\times^{b}$ \\
\midrule
\multicolumn{6}{@{}l}{Training data of the detector} & \makecell{TUSZ \texttt{train}\\+ Siena} & \makecell{TUSZ \texttt{train}\\+ \texttt{dev} (ours)} & CHB-MIT & \makecell{TUSZ \texttt{train}\\+ \texttt{dev}$^{c}$} \\
\bottomrule
\end{tabular}}

\vspace{3pt}
\begin{minipage}{\textwidth}\footnotesize
$^{a}$ Corpus is in the detector's training data. $^{b}$ Detector needs unipolar input, which cannot be recovered from a bipolar montage (Section~\ref{sec:data}). $^{c}$ Split stated by the maintainers of the released weights (SzCORE platform card). $^{d}$ 24 cases; \texttt{chb21} is a later session of \texttt{chb01}.
\end{minipage}
\end{table*}

\subsection{Detectors and the adapter interface}
\label{sec:models}

A detector is integrated through an adapter with two functions: \texttt{detect(eeg, data)} returns a per-sample mask for an in-memory array, and the optional \texttt{compute\_agreement} returns the MCD agreement from the same inputs. The adapter declares the input montage; the framework converts the montage and leaves all other preprocessing to the detector. We audited four detectors that differ in architecture, size, input, preprocessing and output (Table~\ref{tab:models}): SeizureTransformer~\cite{wu2025seizuretransformer}, a U-Net with a transformer bottleneck that ranked first in the 2025 SzCORE challenge; EEGNet~\cite{lawhern2018eegnet}, a compact CNN that we trained on 569 TUSZ \texttt{train}+\texttt{dev} subjects with a subject-wise validation split for early stopping and threshold selection (Supplementary Section~S3); EEGWaveNet~\cite{thuwajit2022eegwavenet}, a five-fold multiscale CNN ensemble; and EventNet~\cite{seeuws2024eventnet}, which predicts event centers and durations instead of thresholding a per-sample score. For the released detectors, the adapters reproduce the upstream predictions exactly on clean data.

\begin{table*}[!htbp]
\centering
\caption{Audited detectors. Preprocessing is the detector's own and is applied after the \RS{} shift. MCD: Monte Carlo dropout available. Attack target: differentiable output attacked by FGSM; decisions use the detector's own post-processing.}
\label{tab:models}
\footnotesize\setlength{\tabcolsep}{4pt}
\begin{tabularx}{\textwidth}{@{}L{2.15cm} L{1.95cm} r L{1.35cm} X L{2.1cm} c L{1.7cm}@{}}
\toprule
Detector & Architecture & Params & Input & Own preprocessing & Output & MCD & Attack target \\
\midrule
Seizure\-Transformer & U-Net + transformer & 44.1M & 19-ch unipolar & per-channel z-score per recording; 0.5--120\,Hz band-pass; 1 and 60\,Hz notches; 60\,s windows & per-sample probability, threshold 0.8, morphological filtering & \checkmark & per-sample probability \\
EEGNet (trained here) & depthwise-separable CNN & 1,649 & 18-ch bipolar & 0.5--40\,Hz band-pass; 128\,Hz; per-channel z-score per recording; 4\,s windows, 1\,s stride & window probability averaged per second, threshold 0.9, gap closing, events $\ge$2\,s & \checkmark & window probability \\
EEGWaveNet & multiscale CNN, 5-fold ensemble & 5$\times$45.7k & 18-ch bipolar & 64\,Hz low-pass; no high-pass; no normalization; 4\,s windows, 1\,s stride & per-fold argmax, majority vote & -- & mean ensemble probability \\
EventNet & event U-Net & 1.66M & 19-ch unipolar & none: raw \textmu V; 120\,s windows & event center and duration maps; peak finding & -- & event-center map \\
\bottomrule
\end{tabularx}
\end{table*}

\subsection{Perturbation library}
\label{sec:shifts}

Each perturbation models a known clinical cause and is applied per channel $c$ to the acquired signal $x_c(t)$ (Table~\ref{tab:shifts}; examples in Fig.~\ref{fig:gallery}; full definitions in Supplementary Section~S4). Amplitudes are physical (\textmu V) or relative to the channel's root mean square $r_c$ or standard deviation $s_c$, so that severity means the same on every corpus. Random components are seeded.

\begin{table*}[!htbp]
\centering
\caption{Perturbation library: 17 environment, 13 noise and 6 adversarial settings, plus the clean baseline (37 settings). $r_c$, $s_c$: per-channel RMS and standard deviation; $\mathcal{N}$: Gaussian; $\mathrm{LP}_{1\,\mathrm{Hz}}$: second-order Butterworth low-pass.}
\label{tab:shifts}
\footnotesize
\begin{tabularx}{\textwidth}{@{}L{2.4cm} L{3.7cm} X L{3.2cm}@{}}
\toprule
Family & Clinical cause & Signal model & Settings \\
\midrule
\multicolumn{4}{@{}l}{\emph{Environment (acquisition)}}\\
Band-pass filter & amplifier/export filter settings & third-order causal Butterworth band-pass $[f_l, f_h]$ & 0.5--40, 1--40, 0.5--25\,Hz \\
Electrode loss & detached or bridged electrode & all channels involving the electrode set to zero in the acquisition montage & T7+T8; O1+O2 \\
Sampling round-trip & low-rate device or export & FFT resampling to $f'$ and back: removes content above $f'/2$ & $f'$ = 50, 100, 200\,Hz \\
ADC quantization & low-resolution converter & uniform quantizer, $2^{b-1}-1$ levels per polarity, full scale = channel maximum or $\pm$100\,\textmu V with clipping & 5, 12 bit; 10 bit at $\pm$100\,\textmu V \\
Impedance noise & poor or drying contact & $x_c + A\,\nu_c/\max|\nu_c|$, $\nu_c = \mathrm{LP}_{1\,\mathrm{Hz}}(\mathcal{N}(0,1))$ & $A$ = 50, 100, 200\,\textmu V \\
Baseline drift & sweat, movement, electrode polarization & 0.2\,Hz sine of 50\,\textmu V; random walk of 0.05--0.5\,Hz noise; piecewise-linear (8 knots); walk and piecewise scaled to $0.1\,s_c$ & 3 modes \\
\multicolumn{4}{@{}l}{\emph{Noise and artifact}}\\
White noise (AWGN) & amplifier and environmental noise & $x_c + n_c$, $n_c\sim\mathcal{N}(0,\,r_c^2\,10^{-\mathrm{SNR}/10})$ & SNR = 50, 20, 5, 2, 0\,dB \\
Mains hum & power-line interference & $x_c + \alpha\, r_c \sin(2\pi\,60\,t + \varphi_c)$ & $\alpha$ = 0.1, 0.5, 1.0 \\
Impulses & electrode pops, movement spikes & spikes of $\pm4\,r_c$ at Bernoulli($p$) samples & $p$ = 0.05, 0.5, 1\% \\
Notch filter & regional notch setting & IIR notch, $Q=30$ & 50\,Hz \\
Mains/notch mismatch & notch set for the wrong grid & 60\,Hz hum ($\alpha=0.1$) followed by a 50\,Hz notch & 1 \\
\multicolumn{4}{@{}l}{\emph{Adversarial (white-box)}}\\
FGSM increase / decrease & worst-case margin probe & one signed-gradient step on the detector's normalized input, Eq.~\eqref{eq:fgsm} & $\varepsilon$ = 0.01, 0.05, 0.10 \\
\bottomrule
\end{tabularx}
\end{table*}

\paragraph{Adversarial probe} Let $u=\rho(x)$ be a window of the detector's normalized input, $p=h(u)$ its seizure probability, $y^\star$ the target ($y^\star=1$, `increase', provokes detections; $y^\star=0$, `decrease', suppresses them) and $\mathcal{L}=\mathrm{BCE}(p,y^\star)$ the binary cross-entropy. Minimizing the first-order expansion of $\mathcal{L}$ over the box $|\delta_{c,t}|\le\varepsilon\,\sigma_c$, with $\sigma_c$ the per-window standard deviation of channel $c$, gives the fast gradient sign method (FGSM)~\cite{goodfellow2015fgsm}:
\begin{equation}
u' = u - \varepsilon\,\sigma \odot \operatorname{sign}\big(\nabla_u \mathcal{L}\big),
\qquad
\mathcal{L}(u')-\mathcal{L}(u) \approx -\,\varepsilon \textstyle\sum_{c,t}\sigma_c\,|\partial_{u_{c,t}}\mathcal{L}| .
\label{eq:fgsm}
\end{equation}
Scaling by $\sigma_c$ makes $\varepsilon$ dimensionless and comparable across detectors: $\varepsilon=0.05$ perturbs each channel by 5\% of its standard deviation. A single first-order step gives a \emph{lower bound} on worst-case vulnerability.

\begin{figure*}[!htbp]
\centering
\includegraphics[width=\textwidth]{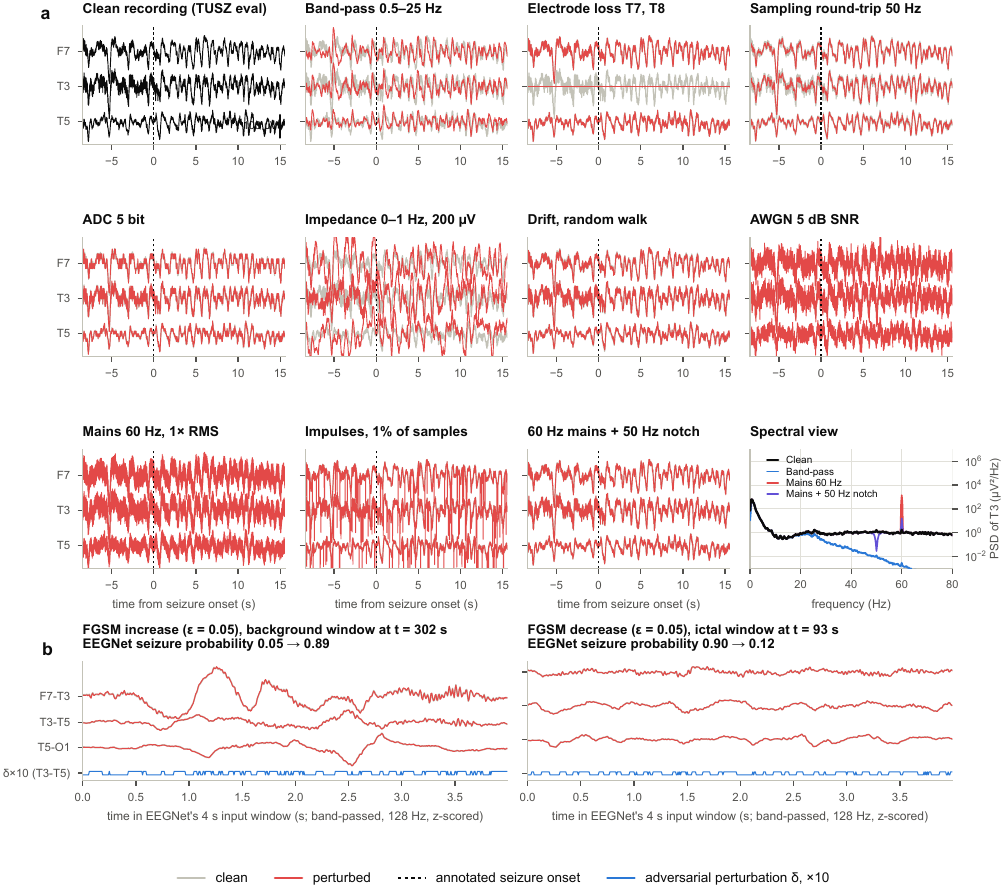}
\caption{The perturbation library applied to a TUSZ \texttt{eval} recording around an annotated seizure onset (dashed line). (a) Ten shifts produced by the benchmark's own code with the settings of Table~\ref{tab:shifts}; gray: clean, red: perturbed. The spectral panel shows the T3 power spectral density for the clean recording, the 0.5--25\,Hz band-pass, 60\,Hz mains and the mains/notch mismatch. (b) FGSM ($\varepsilon=0.05$) in EEGNet's normalized input space: one step moves a background window from seizure probability 0.05 to 0.89, and an ictal window from 0.90 to 0.12, while the perturbed traces overlap the clean ones almost exactly; the perturbation $\delta$ is magnified ten-fold.}
\label{fig:gallery}
\end{figure*}

\FloatBarrier
\subsection{Scoring}
\label{sec:scoring}

\begin{figure*}[!htbp]
\centering
\includegraphics[width=\textwidth]{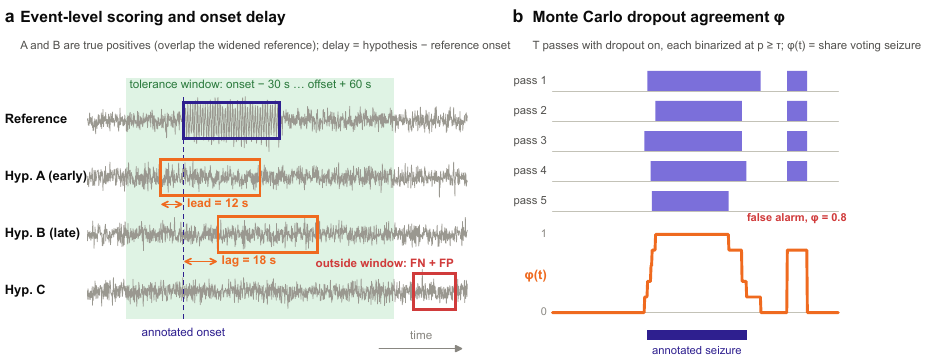}
\caption{Scoring and uncertainty. (a) SzCORE event scoring. A hypothesis is a true positive if it overlaps the reference seizure widened by 30\,s before onset and 60\,s after offset (green). Onset delay is the hypothesis onset minus the annotated onset: negative is a lead (A), positive a lag (B). A hypothesis outside the window (C) is a false positive and leaves the seizure missed. (b) Monte Carlo dropout agreement: $T$ stochastic passes with dropout active are each binarized at $p\ge\tau$, and $\varphi(t)$ is the share of passes voting seizure. A false alarm repeated in four of five passes has $\varphi=0.8$: it is confident, not uncertain.}
\label{fig:scoring}
\end{figure*}

Outputs were scored with the SzCORE \texttt{timescoring} library~\cite{dan2024szcore} (Fig.~\ref{fig:scoring}a). In event-based scoring, a hypothesis event is a true positive (TP) if it overlaps a reference event widened by 30\,s before onset and 60\,s after offset; other hypothesis events are false positives (FP). With $N_{\mathrm{ref}}$ reference events and total duration $D$,
\begin{equation}
\begin{gathered}
\Se=\mathrm{TP}/N_{\mathrm{ref}},\qquad
\PPV=\mathrm{TP}/(\mathrm{TP}+\mathrm{FP}),\\
\Fone=\frac{2\,\Se\cdot\PPV}{\Se+\PPV},\qquad
\FPd=\mathrm{FP}/(D/86400\,\mathrm{s}) .
\end{gathered}
\label{eq:metrics}
\end{equation}
Counts are pooled over a corpus before the ratios are taken; the SzCORE challenge averages per subject~\cite{dan2025challenge}, so values are not directly comparable to its leaderboard. Event F1 is the primary metric, and FP/24\,h is reported separately because alarm burden decides clinical acceptability~\cite{beniczky2021automated}. Timing matters as much as detection: a convulsive seizure lasting 5\,min is treated as status epilepticus~\cite{trinka2015se}, and delayed first-line treatment worsens outcome~\cite{gainzalein2018time}, so a late alarm consumes the treatment window even when F1 counts the seizure as detected. \emph{Onset delay} is $d = t^{\mathrm{hyp}}_{\mathrm{on}}-t^{\mathrm{ref}}_{\mathrm{on}}$ for each detected reference event, with $t^{\mathrm{hyp}}_{\mathrm{on}}$ the onset of the earliest matched hypothesis; $d<0$ is a lead and $d>0$ a lag. Each run reports the number of detected seizures, the share detected early and the median delay, lag and lead.

\subsection{Uncertainty: Monte Carlo dropout agreement}
\label{sec:mcd}

Dropout at test time approximates sampling from a variational posterior over network weights, so the spread of $T$ stochastic passes estimates epistemic uncertainty~\cite{gal2016dropout,kendall2017uncertainties}. For the two detectors with dropout layers (SeizureTransformer, EEGNet), \RS{} keeps dropout active, holds normalization layers in inference mode, binarizes each pass $p^{(i)}(t)$ at $\tau=0.8$ and computes the agreement index
\begin{equation}
\varphi(t)=\frac{1}{T}\sum_{i=1}^{T}\mathbb{1}\big[p^{(i)}(t)\ge\tau\big]\in[0,1],
\label{eq:phi}
\end{equation}
a Monte Carlo estimate of the posterior probability that the sample is called a seizure ($\varphi=0.5$: maximal disagreement). We averaged $\varphi$ over all samples and over seizure, background, false-positive and false-negative samples, and summarized confidence by the certainty $c=|2\varphi-1|$ (1: all passes agree; 0: an even split). A change in $c$ is informative only beside the change in accuracy~\cite{seoni2026calibration,ovadia2019uncertainty}, so each shifted setting received one reading: \emph{honest degradation} if F1 degraded materially and $c$ fell significantly, \emph{overconfident failure} if $c$ rose, \emph{unflagged failure} if $c$ was unchanged, \emph{early warning} if $c$ fell without material loss of F1, and \emph{robust} otherwise. MCD runs used $T=5$ on the first ten subjects of each corpus for all 37 settings, compared with a clean run on the same subset; $\Delta\Fone$ came from the full evaluation.

\subsection{Statistical analysis}
\label{sec:stats}

Recordings of a subject are dependent, so we resampled subjects~\cite{efron1993bootstrap,field2007bootstrap}: $B=2000$ bootstrap samples per pair and scope, with all 37 runs recomputed on the same resampled subjects. $\Delta\Fone$, $\Delta\Se$ and the FP/24\,h ratio are therefore paired contrasts, and their percentile 95\% confidence intervals (CIs) reflect between-subject variability of the effect. A shifted setting is a \emph{material degradation} when the CI of $\Delta\Fone$ lies below zero and $\Delta\Fone\le-0.02$. Onset statistics and the MCD certainty were bootstrapped in the same paired way, the latter on the subset subjects. Agreement between robustness profiles is the Spearman correlation $\rho$ of $\Delta\Fone$ over the 36 shifted settings. The analysis is descriptive, without multiplicity correction (algorithm in Supplementary Section~S5).

\subsection{Implementation}
\label{sec:impl}

\RS{} is written in Python and PyTorch. One command runs, scores and logs a setting, and a resumable queue runs the grid in parallel. The audit comprised 444 runs: 296 full-evaluation runs (37 settings $\times$ 8 pairs) and 148 MCD runs (37 $\times$ 4 pairs). All runs were executed uniformly on CPU with seeded random components, taking about 590 CPU job-hours (Supplementary Section~S9).

\FloatBarrier
\section{Results}
\label{sec:results}

\subsection{Clean held-out performance}

\begin{figure*}[!htbp]
\centering
\includegraphics[width=\textwidth]{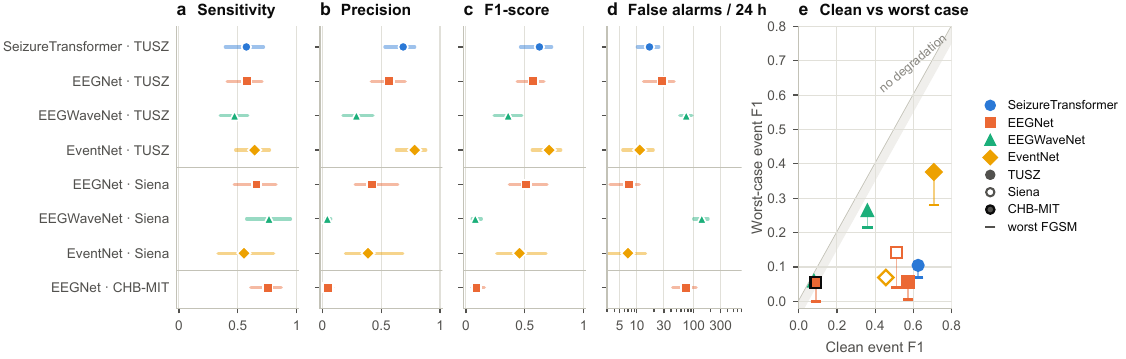}
\caption{Clean held-out performance. (a--d) Event sensitivity, precision, F1 and false alarms per 24\,h (log scale) for each valid detector--corpus pair; markers: point estimates; bars: subject-level bootstrap 95\% CIs. (e) Clean event F1 against the worst F1 over the 30 non-adversarial shifts (marker) and over the six FGSM settings (bar end); distance below the diagonal is the worst-case loss, and pairs with low clean F1 sit near it because F1 cannot fall below zero (floor effect).}
\label{fig:baselines}
\end{figure*}

Clean performance varied more between corpora than between detectors (Fig.~\ref{fig:baselines}, Table~\ref{tab:baselines}). On TUSZ, event F1 ranged from 0.36 (EEGWaveNet) to 0.71 (EventNet, also the fewest false alarms: 11.3 per 24\,h), and EEGNet, with 1,649 parameters, came within 0.05 of the 44.1M-parameter SeizureTransformer. EventNet fell to 0.46 on Siena, and the two detectors tested far from their training data, EEGWaveNet on Siena and EEGNet on CHB-MIT, kept sensitivity of 0.77 and 0.76 at 140 and 74 false alarms per day (F1 0.08 and 0.09).

\begin{table*}[!htbp]
\centering
\caption{Clean held-out performance (full evaluation), with subject-level bootstrap 95\% CIs. TP/ref: detected over reference seizure events. Sample F1: sample-based scoring at 1\,Hz.}
\label{tab:baselines}
\footnotesize\setlength{\tabcolsep}{3pt}
\resizebox{\textwidth}{!}{%
\begin{tabular}{@{}l l r r c c c c r@{}}
\toprule
Detector & Corpus & Subj. & TP/ref & Sensitivity & Precision & Event F1 & FP/24\,h & Sample F1 \\
\midrule
\csname @@input\endcsname tables/tab_baselines.tex
\bottomrule
\end{tabular}}
\end{table*}

\FloatBarrier
\subsection{The robustness landscape}

\begin{figure*}[!htbp]
\centering
\includegraphics[width=\textwidth]{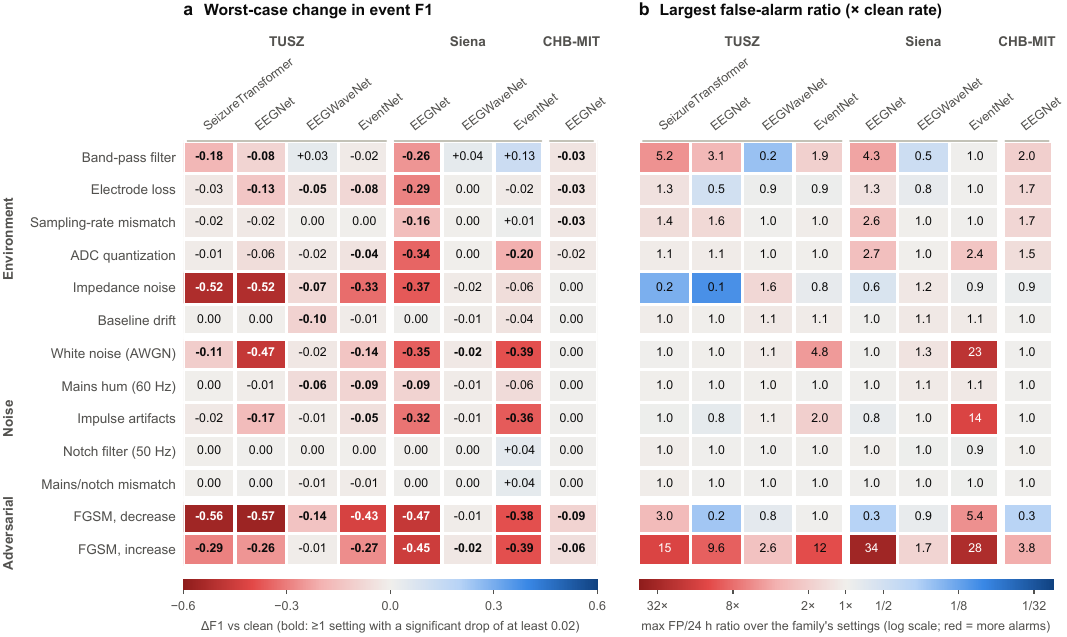}
\caption{Robustness landscape (full evaluation). (a) Worst change in event F1 over the settings of each shift family, relative to the pair's clean run; bold: at least one significant drop of at least 0.02 (paired bootstrap). (b) Largest ratio of false alarms per 24\,h to the clean rate over the family's settings; red: alarm flooding, blue: suppression. Columns are grouped by corpus.}
\label{fig:landscape}
\end{figure*}

Fig.~\ref{fig:landscape} maps the worst change in event F1 and in false alarms per shift family and pair; Table~\ref{tab:summary} counts material degradations. Of 288 shifted settings, 89 were material degradations: 34 of 136 environment, 23 of 104 noise and 32 of 48 adversarial settings. Six settings changed no pair by 0.02 F1 or more, and notch filtering changed none by more than 0.04, so the library separates harmless from harmful conditions. Robustness was not predicted by clean accuracy (rank correlation of clean and worst non-adversarial F1, $\rho=0.50$): the most accurate detector on TUSZ, EventNet, still lost 0.33 F1 to impedance noise and 0.43 to FGSM. The pairs that lost least had the lowest clean F1, the floor effect of Section~\ref{sec:formal}; some shifts even improved them by suppressing false alarms (nine settings for EEGNet on CHB-MIT).

\begin{table*}[!tb]
\centering
\caption{Robustness summary per detector--corpus pair (full evaluation). Material degradations: settings whose paired 95\% CI of $\Delta$F1 lies below zero with $\Delta$F1 $\le -0.02$, out of the family's settings. Worst F1 over the 30 non-adversarial settings (with the setting) and over the 6 FGSM settings. Max FP ratio: largest FP/24\,h relative to the clean run.}
\label{tab:summary}
\footnotesize\setlength{\tabcolsep}{3pt}
\resizebox{\textwidth}{!}{%
\begin{tabular}{@{}l l c c c c l c c@{}}
\toprule
 & & & \multicolumn{3}{c}{Material degradations} & & & \\
\cmidrule(lr){4-6}
Detector & Corpus & Clean F1 & Env. & Noise & Adv. & \makecell{Worst F1,\\non-adversarial} & \makecell{Worst F1,\\FGSM} & \makecell{Max FP\\ratio} \\
\midrule
\csname @@input\endcsname tables/tab_summary.tex
\bottomrule
\end{tabular}}
\end{table*}

\subsection{Severity--response curves}

\begin{figure*}[!tb]
\centering
\includegraphics[width=\textwidth]{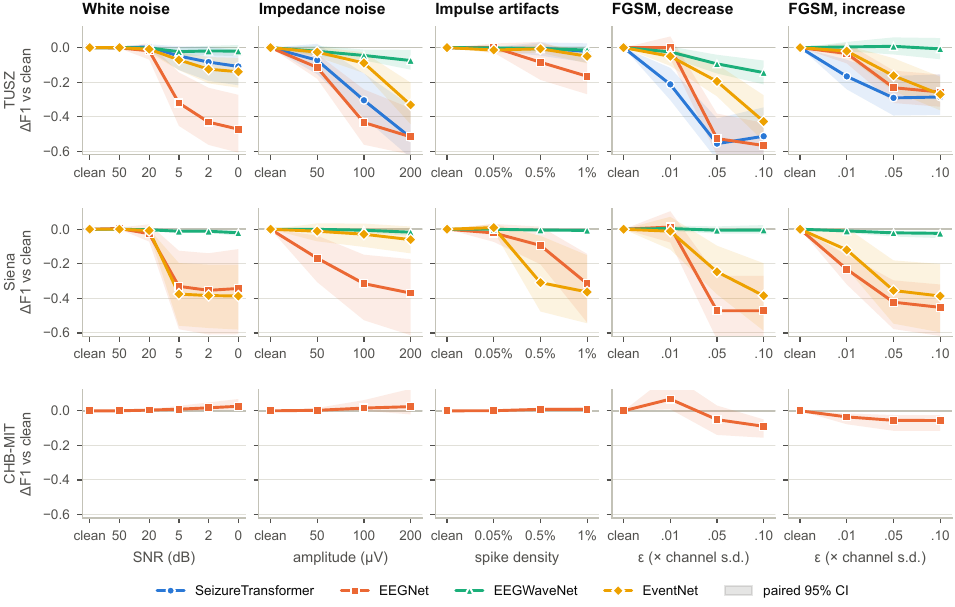}
\caption{Severity--response curves of event F1 (change relative to the clean run, with paired bootstrap 95\% CI bands) for five families, by corpus; severity increases to the right. FGSM $\varepsilon$ is relative to the per-window channel standard deviation of each detector's normalized input. Sensitivity curves are in Supplementary Fig.~S2.}
\label{fig:curves}
\end{figure*}

Degradation grew with severity, with characteristic shapes (Fig.~\ref{fig:curves}). White noise cost at most 0.03 F1 at 20\,dB SNR and above, then dropped abruptly between 20 and 5\,dB for EEGNet on both corpora and for EventNet on Siena. Impedance noise degraded the z-scoring detectors progressively. Adversarial curves were the steepest: FGSM at $\varepsilon=0.05$ reduced F1 by 0.56 (SeizureTransformer) and 0.53 (EEGNet) on TUSZ. `Decrease' suppressed nearly all detections (EEGNet sensitivity 0.58 to 0.02); `increase' flooded alarms, up to 14.5 times the clean rate for SeizureTransformer and 34 times for EEGNet on Siena. EEGWaveNet was least affected ($\Delta\Fone\ge-0.14$), consistent with its low clean F1 and with gradient masking by its non-differentiable majority vote~\cite{athalye2018obfuscated}.

\subsection{Two failure modes: silent failure and alarm flooding}

\begin{figure*}[!htbp]
\centering
\includegraphics[width=\textwidth]{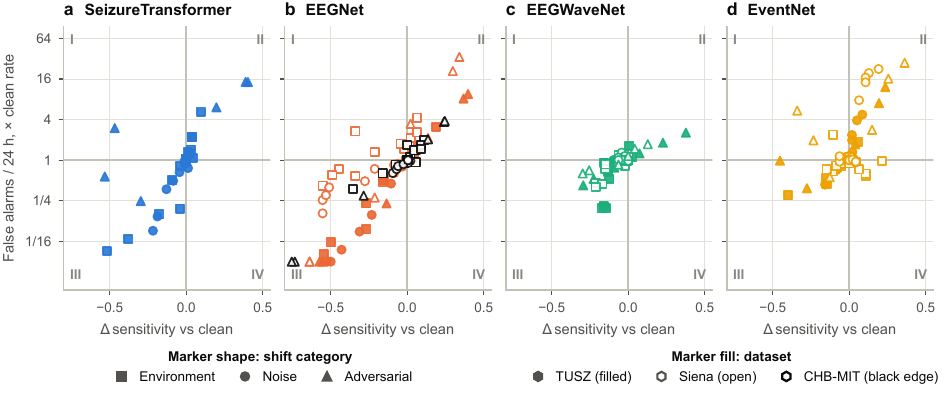}
\caption{Failure modes. Each marker is one shifted setting of one valid pair (full evaluation): change in event sensitivity against the false-alarm ratio (log scale). Quadrant I: missed seizures and more alarms; II: more detections and more alarms (alarm flooding); III: missed seizures with fewer alarms (silent failure); IV: benign. Marker shape: family; fill: corpus.}
\label{fig:failure}
\end{figure*}

Plotting the change in sensitivity against the false-alarm ratio separates two clinically different failure modes (Fig.~\ref{fig:failure}). In \emph{silent failure} (quadrant III) the detector misses seizures and raises fewer alarms, so nothing signals the problem; in \emph{alarm flooding} (quadrants I--II) false alarms multiply.

Impedance noise produced the clearest silent failures: at 200\,\textmu V on TUSZ, sensitivity fell from 0.57 to 0.06 (SeizureTransformer) and from 0.58 to 0.03 (EEGNet), with false alarms at 4\% of the clean rate. Both detectors normalize each channel by its standard deviation over the recording. With independent noise $n$ added to the EEG $s$, the network receives the EEG scaled by
\begin{equation}
g=\mathrm{sd}(s)/\mathrm{sd}(s+n)=\big(1+\sigma_n^2/\sigma_s^2\big)^{-1/2},
\label{eq:gain}
\end{equation}
computed on the signal the detector normalizes. Later filters remove most of the slow noise, but it has already inflated the divisor. On twelve TUSZ recordings, the median gain at 50, 100 and 200\,\textmu V was 0.74, 0.49 and 0.27 for SeizureTransformer and 0.51, 0.28 and 0.15 for EEGNet, which lost more sensitivity at every amplitude (Supplementary Section~S6). EventNet, which does not normalize, was also suppressed (sensitivity 0.64 to 0.24), so large slow components are harmful beyond normalization; EEGWaveNet kept its sensitivity.

Alarm flooding came from FGSM `increase' in every pair and, for EventNet on Siena, from broadband noise: EventNet takes raw \textmu V without filtering or normalization, and on Siena 5\,dB AWGN raised its false alarms 17-fold and 1\% impulses 14-fold, whereas on TUSZ the same noise raised them at most 4.8-fold.

\subsection{Preprocessing determines what a detector absorbs}

Several results follow from each detector's preprocessing (Table~\ref{tab:models}). Slow sine drift materially affected only EEGWaveNet ($\Delta\Fone=-0.10$ on TUSZ), the one filtered detector without a high-pass stage. EventNet, without notch or low-pass, was the only detector hurt by 60\,Hz mains (sensitivity 0.64 to 0.49 at amplitude 1.0). An additional band-pass \emph{improved} the two detectors that do not band-limit from below: on Siena a 0.5--40\,Hz filter raised EventNet's F1 from 0.46 to 0.61 and cut its false alarms by 41\%, and EEGWaveNet's false alarms fell by 54--81\% on both corpora. Part of their cross-corpus gap is thus out-of-band content. For detectors that already band-pass, the stacked filter cost up to 0.26 F1.

\subsection{Profiles transfer across corpora, not across detectors}
\label{sec:transfer}

For EEGNet and EventNet, the ranking of the 36 shifts by $\Delta\Fone$ was largely preserved from TUSZ to Siena ($\rho=0.86$ and 0.74), less so for EEGWaveNet ($\rho=0.42$), and EEGNet's TUSZ profile did not predict its CHB-MIT profile ($\rho=-0.04$), where the floor effect dominates. On TUSZ, profiles of different detectors agreed only moderately ($\rho$ 0.15--0.71). \emph{Which} shifts hurt is largely a property of the detector and its preprocessing; \emph{how much} depends on the corpus.

\subsection{Onset timing}

\begin{figure*}[!htbp]
\centering
\includegraphics[width=\textwidth]{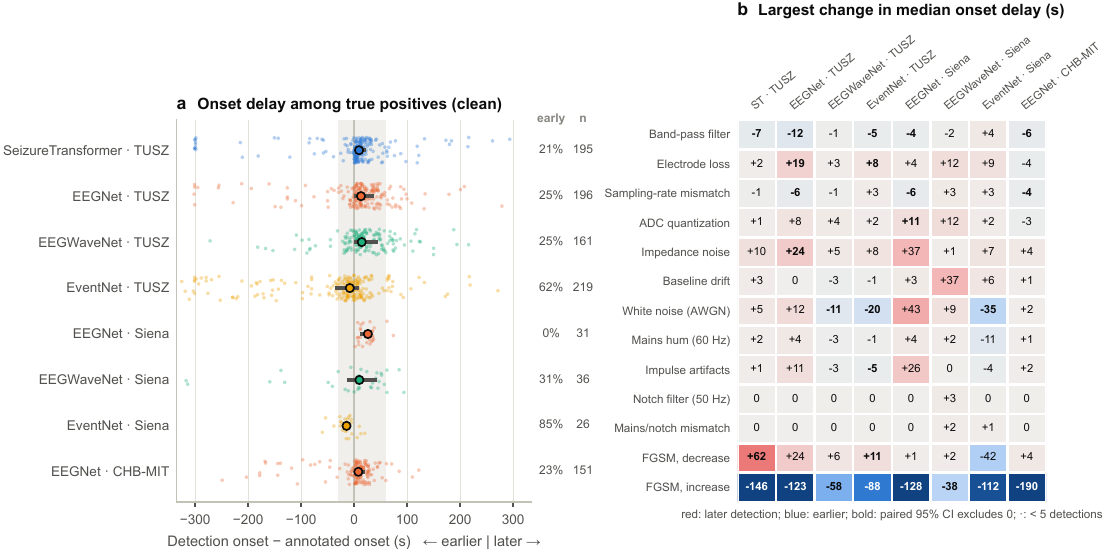}
\caption{Onset timing among event-level true positives (full evaluation). (a) Detection onset minus annotated onset on clean data (points: seizures; bar: interquartile range; circle: median); right: share of early detections and number of true positives. (b) Largest change in median onset delay over the settings of each shift family (shifted minus clean, seconds; red: later, blue: earlier); bold: paired subject-bootstrap 95\% CI excludes zero; dot: fewer than five detections. Changes in all onset statistics, with CIs, are in Supplementary Section~S8.}
\label{fig:onset}
\end{figure*}

On clean data, SeizureTransformer, EEGNet and EEGWaveNet mostly detected after the annotated onset (median delay 8--26\,s; 0--31\% early), whereas EventNet, which places events around predicted centers, fired early in 62\% of TUSZ and 85\% of Siena detections (median $-8$ and $-14$\,s; Fig.~\ref{fig:onset}a).

Shifts moved timing in two clinically opposite directions (Fig.~\ref{fig:onset}b; every statistic with paired CIs in Supplementary Section~S8). Silent-failure shifts also delayed the seizures still detected: the median delay rose significantly in 12 settings, by up to 62\,s for SeizureTransformer under FGSM `decrease' (9 to 71\,s) and by 24\,s for EEGNet under 200\,\textmu V impedance noise, which left 10 of 196 seizures detected. A minute of added delay is a fifth of the five-minute window after which a convulsive seizure is treated as status epilepticus~\cite{trinka2015se}. Alarm flooding moved timing the other way: under FGSM `increase' the share of early detections rose in 18 of 24 settings, to as much as 96\% (EEGNet on CHB-MIT, from 23\%), with median leads of up to 3\,min. F1 credits these alarms as hits; timing shows that they carry no onset information.

\subsection{Uncertainty read with accuracy}

\begin{figure*}[!htbp]
\centering
\includegraphics[width=\textwidth]{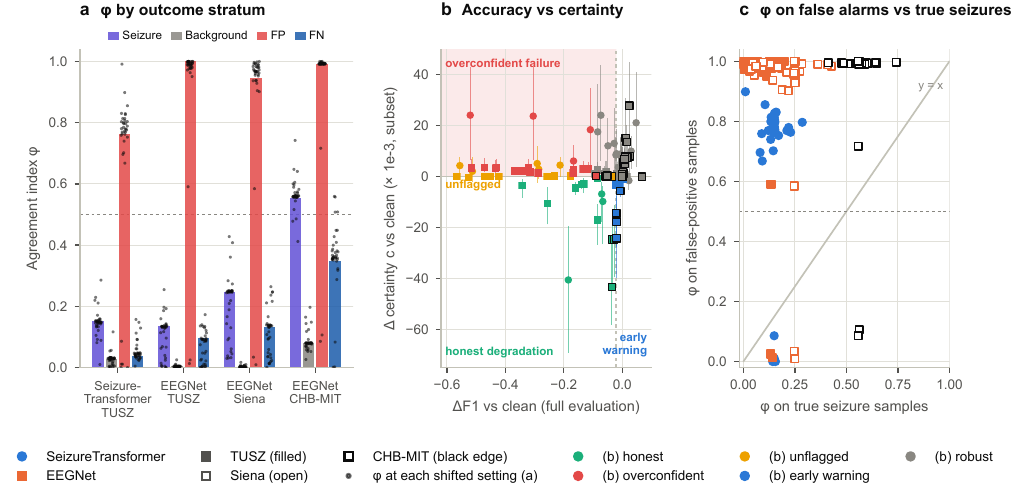}
\caption{Monte Carlo dropout agreement $\varphi$ (ten-subject research subsets, $T=5$). (a) Mean $\varphi$ by stratum on clean data (bars) and at each of the 36 shifted settings (dots); FP, FN: false-positive and false-negative samples; dashed line: maximal disagreement. (b) Change in certainty $c=|2\varphi-1|$ (subset, paired 95\% CI) against change in event F1 (full evaluation) for every shifted setting, colored by reading; dashed line: material-degradation threshold; shaded: overconfident failure. (c) $\varphi$ on false-positive against seizure samples for every setting: natural false alarms are at least as confident as true detections, whereas FGSM-injected alarms (bottom) are not repeated by the stochastic passes.}
\label{fig:uncertainty}
\end{figure*}

On clean data, $\varphi$ on false-positive samples was 0.76 for SeizureTransformer and 0.95--1.00 for EEGNet (Fig.~\ref{fig:uncertainty}a): natural false alarms were repeated by nearly every stochastic pass. Under shift, certainty moved by only $10^{-3}$--$10^{-2}$, yet paired CIs resolved these changes, and beside $\Delta\Fone$ their direction separated the failure modes (Table~\ref{tab:mcd}, Fig.~\ref{fig:uncertainty}b). Filtering shifts (band-pass, sampling round-trip, coarse ADC) lowered certainty: 13 degradations were honest (SeizureTransformer, 1--40\,Hz: $\Delta\Fone=-0.18$, $\Delta c=-0.041$ [$-0.069$, $-0.019$]), and in eight settings certainty fell before F1 did, an early warning. Additive and impedance noise, impulses and electrode loss raised certainty as detection collapsed, giving 23 overconfident failures (EEGNet on TUSZ at 200\,\textmu V: $\Delta\Fone=-0.52$, $\Delta c=+0.003$ [$+0.001$, $+0.007$]); certainty on seizure samples rose by up to 0.07: the detector grew surer that seizures were background. FGSM left overall certainty unchanged (16 of 17 unflagged failures), but its injected alarms were rarely repeated by the stochastic passes: $\varphi$ on false positives fell to 0.01--0.11 at $\varepsilon\ge0.05$ (Fig.~\ref{fig:uncertainty}c). Only 13 of the 53 degradations lowered certainty.

\begin{table*}[!htbp]
\centering
\caption{Accuracy and certainty read together. For every setting and pair: $\Delta$F1, the change in event F1 from the clean run (full evaluation; bold: material degradation), and $\Delta c$, the change in Monte Carlo dropout certainty $c=|2\varphi-1|$ averaged over all samples, in units of $10^{-3}$. $\Delta c$ comes from the ten-subject MCD subsets ($T=5$) and is compared with the clean run of the same subset; despite their size these runs resolve small changes, and all paired 95\% CIs are listed in Supplementary Section~S8. Shading gives the reading: \colorbox{rdH}{H} honest degradation (certainty fell significantly with a material F1 loss), \colorbox{rdO}{O} overconfident failure (certainty rose), \colorbox{rdU}{U} unflagged failure (unchanged), \colorbox{rdW}{W} early warning (certainty fell without material F1 loss); unshaded: robust. --: EEGWaveNet and EventNet contain no dropout layers, so MCD cannot be applied without altering them.}
\label{tab:mcd}
\scriptsize\setlength{\tabcolsep}{2.2pt}
\resizebox{\textwidth}{!}{%
\begin{tabular}{@{}l rr rr rr rr rr rr rr rr@{}}
\toprule
 & \multicolumn{8}{c}{TUSZ} & \multicolumn{6}{c}{Siena} & \multicolumn{2}{c}{CHB-MIT} \\
\cmidrule(lr){2-9}\cmidrule(lr){10-15}\cmidrule(l){16-17}
 & \multicolumn{2}{c}{SeizureTransformer} & \multicolumn{2}{c}{EEGNet} & \multicolumn{2}{c}{EEGWaveNet} & \multicolumn{2}{c}{EventNet} & \multicolumn{2}{c}{EEGNet} & \multicolumn{2}{c}{EEGWaveNet} & \multicolumn{2}{c}{EventNet} & \multicolumn{2}{c}{EEGNet} \\
Setting & $\Delta$F1 & $\Delta c$ & $\Delta$F1 & $\Delta c$ & $\Delta$F1 & $\Delta c$ & $\Delta$F1 & $\Delta c$ & $\Delta$F1 & $\Delta c$ & $\Delta$F1 & $\Delta c$ & $\Delta$F1 & $\Delta c$ & $\Delta$F1 & $\Delta c$ \\
\midrule
\csname @@input\endcsname tables/tab_mcd.tex
\bottomrule
\end{tabular}}
\end{table*}

\FloatBarrier
\section{Discussion}
\label{sec:discussion}

\subsection{Principal findings}

Robustness to realistic EEG shifts is a separate axis of detector quality that clean accuracy does not measure. Almost a third of the shifted settings, and two thirds of the adversarial ones, materially degraded detection, in two clinically distinct forms: silent loss of sensitivity and alarm flooding. The failures had mechanisms rooted in preprocessing: per-recording normalization turns slow impedance noise into attenuation of the ictal signal (Eq.~\eqref{eq:gain}), and missing high-pass, notch or low-pass stages expose a detector to drift, mains or out-of-band content. Structured changes of 5\% of the channel standard deviation removed or flooded detections, so decision margins are narrow relative to natural EEG variability. Failures also cost time: silenced detectors caught their remaining seizures up to a minute later, inside a treatment window of minutes. And read with accuracy, dropout certainty fell honestly under filtering shifts but rose under the noise that silenced detection, so most failures did not lower certainty.

Four recommendations follow. (i) Report a detector's robustness profile on every valid held-out corpus, not a single clean score (a checklist is given in Supplementary Section~S10). (ii) Treat preprocessing as part of the model: normalizing after band-limiting, with robust scale estimates, would mitigate the impedance failure. (iii) Test specifically for silent failure and delayed detection, which a ward would perceive as a calmer system. (iv) Report uncertainty beside accuracy for every shift, and validate it before relying on it to defer cases~\cite{ovadia2019uncertainty,seoni2026calibration}.

\subsection{Relation to prior work}

SzCORE standardized data, splits and metrics~\cite{dan2024szcore}, and its challenge quantified the loss on unseen clinical data~\cite{dan2025challenge}; \RS{} reuses the same corpora and scoring and attributes losses to specific, nameable causes. Earlier EEG perturbation studies examined the latent spaces of feature encoders~\cite{wagh2022latent}, foundation-model robustness on classification benchmarks~\cite{sirca2026beyond,kastrati2025eegbench} and test-time adaptation~\cite{lee2026tta}. Our audit scores complete detectors with clinical event and timing metrics on long recordings, across detectors and corpora with paired inference. That clean accuracy does not order robustness mirrors computer vision~\cite{hendrycks2019corruptions}, and the overconfident failures echo the degradation of uncertainty under shift~\cite{ovadia2019uncertainty}.

\subsection{Limitations}
\label{sec:limitations}

The perturbations are synthetic signal models whose parameters come from domain knowledge, not from measured artifact statistics. FGSM bounds adversarial vulnerability from below; iterative attacks~\cite{madry2018pgd} are implemented but were deferred for compute (about eight times FGSM), and gradient masking may flatter EEGWaveNet. Siena contributes 47 seizures, and certainty changes come from subsets of 36--59 reference events with $T=5$ passes. Only two detectors support dropout, and four detectors were audited; foundation models~\cite{jiang2024labram,wang2025cbramod} and other SzCORE entries can be added through the same adapter. EventNet's TUSZ validity rests on its maintainers' statement, and EEGNet's clean performance reflects our training choices. The audit covers scalp 10--20 EEG only.

\subsection{Conclusions}
\label{sec:conclusions}

Seizure detectors that are accurate on clean held-out EEG can fail, silently or loudly, under acquisition changes, artifacts and small structured perturbations, often with confidence and with delayed detection. \RS{} turns this into a reproducible, model-agnostic audit with explicit validity rules, paired statistics and mechanistic interpretation, and it extends naturally to compound shifts and robustness-aware training. We propose that robustness profiles accompany accuracy in the pre-deployment assessment of EEG seizure detectors.

\section*{Code and data availability}
\label{sec:availability}
\RS{} is available at \url{https://github.com/iMohammad97/RobustSeiz}, with the adapters for the four detectors, the trained EEGNet weights, the registry of all 444 runs and the scripts that reproduce the run plan. The corpora are public: TUSZ from the Temple University Hospital EEG resource (registration required), Siena and CHB-MIT from PhysioNet. The analysis scripts, per-recording counts and bootstrap results behind every figure and table are provided with the Supplementary Material.

\section*{Acknowledgements}
We thank the maintainers of TUSZ, Siena and CHB-MIT, the SzCORE initiative for the open benchmark and the reproductions of EEGWaveNet and EventNet, and the authors of SeizureTransformer for releasing their weights.

\paragraph{Ethical approval} This study is a secondary analysis of publicly available, de-identified EEG corpora (TUSZ, Siena Scalp EEG Database, CHB-MIT), used under their data use agreements. No new data were collected, no participants were contacted and no identifiable information was accessed, so no additional ethical approval or informed consent was required; approval and consent for the original recordings were obtained by the data providers.

\paragraph{Funding} This research did not receive any specific grant from funding agencies in the public, commercial, or not-for-profit sectors. This work was carried out entirely in the authors' own personal capacity.

\paragraph{Declaration of competing interest} The authors declare that they have no known competing financial interests or personal relationships that could have appeared to influence the work reported in this paper.

\paragraph{CRediT authorship contribution statement} \textbf{Mohammad Mohammadi:} Conceptualization, Methodology, Software, Validation, Formal analysis, Investigation, Data curation, Writing -- original draft, Writing -- review \& editing, Visualization. \textbf{Alireza Zarei:} Conceptualization, Supervision, Writing -- review \& editing.

\paragraph{Supplementary material} Supplementary material associated with this article (proofs, full perturbation definitions, statistical algorithm, complete per-setting results for every pair and scope, onset and uncertainty tables, computational cost, a proposed robustness report and reproduction instructions) is provided in a separate file.

\section*{Declaration of generative AI and AI-assisted technologies in the manuscript preparation process}
Claude (Anthropic) was used in the drafting of this manuscript. All changes made with AI were reviewed and edited by the authors as needed, and the authors take full responsibility for the content of the published article.

\bibliographystyle{elsarticle-num}
\bibliography{references}

\clearpage
\pdfximage{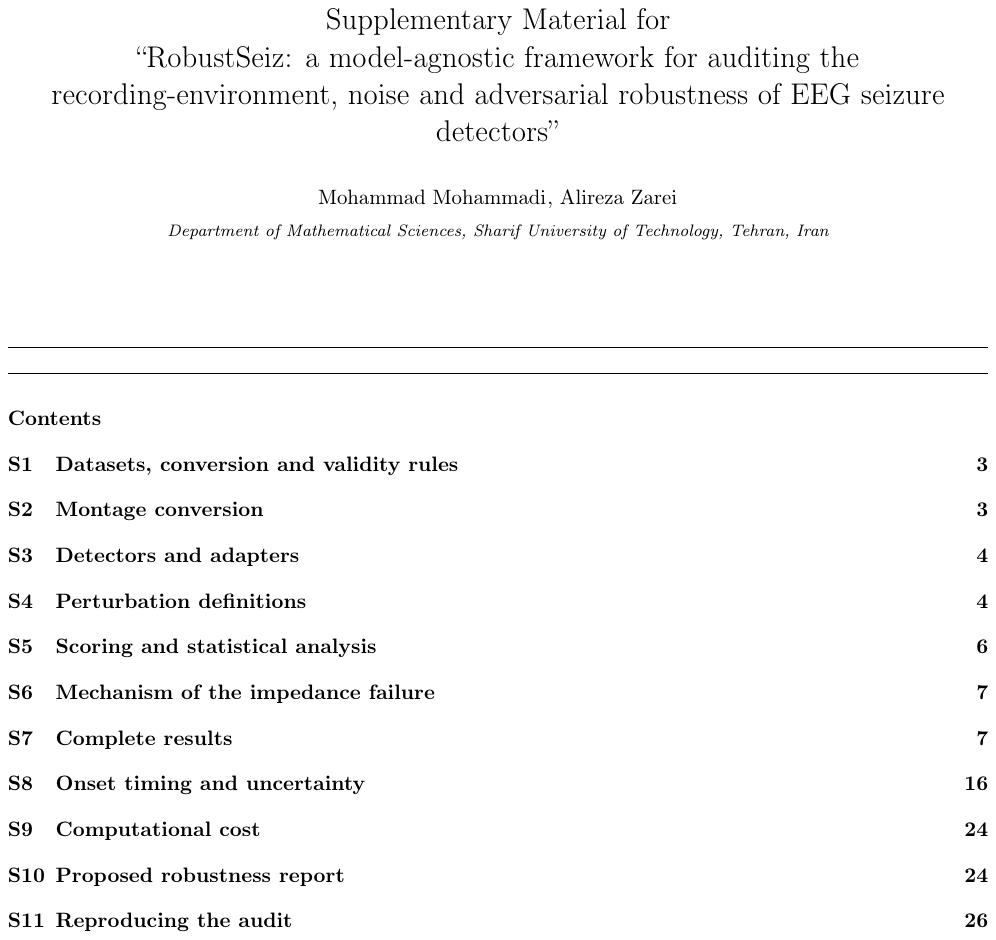}
\newcount\suppN \suppN=\pdflastximagepages
\newcount\suppI \suppI=0
\loop\ifnum\suppI<\suppN
  \advance\suppI by 1
  \thispagestyle{empty}
  \edef\suppShip{\noexpand\AddToHookNext{shipout/background}{\noexpand\put(0pt,-\noexpand\paperheight){\noexpand\includegraphics[page=\the\suppI,width=\noexpand\paperwidth,height=\noexpand\paperheight]{supplementary.pdf}}}}
  \suppShip
  \null\clearpage
\repeat

\end{document}